\documentclass[letterpaper, 10 pt, conference]{ieeeconf}  

\usepackage{graphicx}

\usepackage{algorithm}
\usepackage{algorithmic}

\usepackage{subfigure}
\usepackage{amsmath}
\usepackage{amssymb}

\newtheorem{Def}{Definition}

\IEEEoverridecommandlockouts                              

\title{\LARGE \bf
Off-Dynamics Inverse Reinforcement Learning from Hetero-Domain
}

\author{Yachen Kang$^{1}$, Jinxin Liu$^{1}$, Xin Cao$^{2}$ and Donglin Wang$^{1}$
\thanks{$^{1}$ School of Engineering,
        Westlake University, China
        {\tt\small \{kangyachen, liujinxin, wangdonglin\}@westlake.edu.cn}}%
\thanks{$^{2}$Department of Automation, 
        Tsinghua University, Beijing, China
        {\tt\small caox19@mails.tsinghua.edu.cn}}%
}

\begin{document}

\maketitle
\thispagestyle{empty}
\pagestyle{empty}


\begin{abstract}
We propose an approach for inverse reinforcement learning from hetero-domain which learns a reward function in the simulator, drawing on the demonstrations from the real world.
The intuition behind the method is that the reward function should not only be oriented to imitate the experts, but should encourage actions adjusted for the dynamics difference between the simulator and the real world.
To achieve this, the widely used GAN-inspired IRL method is adopted, and its discriminator, recognizing policy-generating trajectories, is modified with the quantification of dynamics difference. 
The training process of the discriminator can yield the transferable reward function suitable for simulator dynamics, which can be guaranteed by derivation.
Effectively, our method assigns higher rewards for demonstration trajectories which do not exploit discrepancies between the two domains.
With extensive experiments on continuous control tasks, our method shows its effectiveness and demonstrates its scalability to high-dimensional tasks.
\end{abstract}

\section{Introduction}

In recent years, deep reinforcement learning (DRL) has shown promising results in numerous complex tasks, from games such as Atari \cite{mnih2015human} and Minecraft \cite{oh2016control} to vision-based robotic control \cite{levine2016end}.
However, the burdensome process of reward handcrafting usually hinders wider application of DRL. 
To avoid tedious reward engineering, \textit{Inverse reinforcement learning (IRL)} \cite{russell1998learning, ng2000algorithms} infers the expert's reward function from demonstration.
The resulting reward function could guide the agent to behave in accordance with the expert's demonstration by utilizing a variety of reinforcement learning methods.
GAN-based imitation learning \cite{ho2016generative,finn2016connection} leverages the framework of Generative Adversarial Network (GAN) \cite{goodfellow2020generative} to minimize the discrepancy between the distribution of the demonstration trajectories and policy-generating trajectories.
Adversarial inverse reinforcement learning (AIRL) \cite{fu2017learning} further formalizes the discriminator so that the learned reward function can be invariant to changing dynamics.

The method mentioned above, no matter where the reward function is being used, learns the reward function in the environment where the demonstration is collected, which is often incompatible with the real application scenario.
For example, in robot learning, the demonstration is most likely collected in the real world \cite{zhang2018deep,yu2018one}, since demonstrating in the real world entails less specialized knowledge.
However, when training the agent, the interaction between the robot and the real environment should be reduced to avoid expensive costs and fatal physical damage. 
Therefore, the training process of the agent should be mainly limited to the simulated environment.
This creates an inevitable domain difference between the dynamics of the two environments for obtaining demonstration and agent training, which makes direct imitation difficult.
AIRL is trying to obtain a reward function that can be generalized to different domain policy training. 
But there is no domain gap between the reward function training process and the demonstration. 
Therefore, what this article proposes is a more realistic new problem setting.

In this paper, we study how to imitate expert's demonstration from real environments in simulation environments, which we describe as IRL from hetero-domain. 
Here we define the former as the source domain and the latter as the target domain. 
Please note that this is different from the setting in the sim2real problem, most of which are trying to use known reward functions in the simulator to train a policy that can be used in real environment across domain gap.
In contrast, we mainly focus on the reward function learning process rather than the policy acquisition process after that.

\begin{figure}[t]
    \centering
    \includegraphics[width=0.26\textwidth]{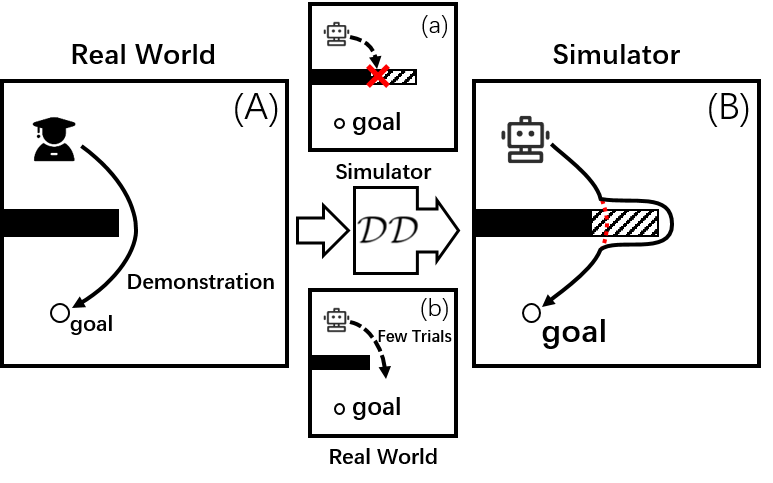}
    \vspace{-10pt}
    \caption{our problem setting}
    \label{Fig1}
    \vspace{-20pt}
\end{figure}

Recent works \cite{liu2019state,Gangwani2020State-only} training imitator only to follow the state trajectories in the expert's demonstration. 
Some describe this paradigm as learning from observations (LfO), trying to directly reproduce the same sequence of states under the dynamics of the target domain to avoid considering the dynamic mismatch between the two domains. 
However, the reliance on the entire trajectory makes this kind of method unsuitable for off-policy RL algorithms.
At the same time, it may be tricky or even impractical to reproduce certain state trajectories on the target domain.
Specifically, as shown in Fig.\ref{Fig1}, the agent learns to get around a wall in the target domain (simulator shown in (B)) based on a demonstration in the source domain (real world shown in (A)). 
The wall in the target domain is longer than that in the source domain. 
In this setting, the elongated wall will hinder the one-step reachability between spatially adjacent states around the corner, inhibiting state-only demonstration trajectories from proceeding in the target domain. 
In other words, when the target domain and the source domain have locally different structures in dynamics, simply mimicking the state may yield poor performance. 
This will be further demonstrated in the subsequent experiments chapter.

To solve this problem, we propose a novel adversarial IRL framework, the Off-Dynamics Inverse Reinforcement Learning (ODIRL).
We imitate the parts that can be achieved in the target domain of the demonstration, and solve the difficult parts by explicitly considering the dynamic differences.
Our framework embeds the quantification of dynamics difference into the process of reward function acquisition.
More specifically, we achieve reward function acquisition through a framework resembling AIRL, but consider dynamics differences in derivation. 
The resulting form of the discriminator in ODIRL explicitly contains a dedicated term, $\mathcal{DD}$ (Dynamics Difference) module, characterizing the dynamics difference between the source and the target domain, which is learned from experience in the target domain and a few trials in the source domain (shown in (a) and (b)).
$\mathcal{DD}$ can be interpreted as a comparison of the transition dynamics in the two domains, and its value is estimated with two binary classifiers.

The main contribution of the paper can be summarized as:
\begin{itemize}
    \item We propose a more realistic problem setting, imitating the real world (hetero-domain) demonstration in the simulator. 
    The real world and the simulator in which the imitator learns the reward function have different dynamics.
    \item We propose a novel adversarial IRL framework, which can solve inverse reinforcement learning from hetero-domain demonstration problem by embedding the quantification of dynamics difference into the process of reward function acquisition.
    \item With extensive experiments on continuous control tasks in the MuJoCo physics engine \cite{todorov2012mujoco}, we demonstrate that our method is effective and scalable in high-dimensional tasks.
\end{itemize}

\section{Related Work}

Behavioral cloning (BC) \cite{bain1995framework} is a straight-forward method to imitate demonstration in a supervised way.   
However, behavioral cloning suffers from the problem of compounding errors as shown by \cite{ross2010efficient}. 
Inverse reinforcement learning (IRL) \cite{russell1998learning,ng2000algorithms,argall2009survey}, on the other hand, aims to first reveal the reward function of the expert, and then use the learned reward function to train the policy.
This kind of method can infer the purpose of expert demonstration to a certain extent, which makes it have stronger generalization and avoids compounding errors.
However, there exists an inherent ambiguity that an expert policy might be explained by several different reward functions.
A current mainstream to solve this ambiguity is to leverage the MaxEnt theories in IRL framework \cite{ziebart2008maximum,ziebart2010modeling}. 

In recent years, the demand for large continuous space applications has promoted the development of GAN-based imitation learning.
Representative methods of this branch include generative adversarial imitation learning (GAIL) \cite{ho2016generative}.
GAIL uses a generator-discriminator adversarial framework to make the state-action distribution of the learned policy as close as possible to the expert demonstration.
However, GAIL does not explicitly recover the reward function, which deters greater transferability in practice. 
GAN-based Guided Cost Learning (GAN-GCL) \cite{finn2016connection} modifies the discriminator to a special form, proves the essential consistency with Guided Cost Learning (GCL) \cite{finn2016guided}, and provides a theoretical framework for interleaving the process of policy improvement and reward acquisition. 
Adversarial inverse reinforcement learning (AIRL) \cite{fu2017learning} moves on to propose a straightforward conversion of discriminator into the single state and action case, and proves that the state-only part of the reward function can generalize to different domains. 
Empowerment-based Adversarial Inverse Reinforcement Learning (EAIRL) \cite{qureshi2018adversarial} formalizes the reward shaping term in the AIRL reward function into empowerment, so that the reward function with action can also be generalized to different domains. 
Probabilistic Embeddings for Meta-Inverse Reinforcement Learning (PEMIRL) \cite{yu2019meta} integrates the priors learned on multiple tasks through meta learning, and can learn few-shot demonstration under new tasks or domains. 
These methods still imitate the demonstration from the same domain as the agent learns in, impeding the application closer to reality. 

Some recent work attempts to solve the problem that arises when there is a domain gap between the demonstration trajectory and the training environment. 
State Alignment based Imitation Learning (SAIL) \cite{liu2019state} tries to solve this problem by imitating state-only demonstrations by employing both local and global alignment to regularize imitator’s policy. 
However, SAIL needs to learn an inverse dynamics model, and the data used to train such model have to be collected separately in the target domain.
As the complexity of the environment increases, this procedure often requires complex exploratory algorithms.
Since it is not directly related to the task goal, the huge workload may not be cost-effective compared to the performance improvement.
Indirect Imitation Learning (I2L) \cite{Gangwani2020State-only} also imitates the state-only demonstrations, which learns a surrogate policy obtained by minimizing Wasserstein distance to the observation between state visitations and then adopts AIRL to imitate from such surrogate policy. 
Like we mentioned in introduction, not all state-only demonstrations can perform normally in the target domain, which could cause imitation failure. 
And the dependence on the entire trajectory makes the off-policy RL algorithm inapplicable, which reduces the data efficiency of these methods. 

Imitation learning with horizon-adaptive inverse dynamics (HIDIL) \cite{jiang2020offline} learns two policies so that the agent can learn to imitate the expert's demonstration by using a misspecified simulator in an offline setting.
However, the lack of learning the reward function limits its application.
Domain Adaptive Imitation Learning (DAIL) \cite{kim2020domain} learns state, action maps between the source and target domain by minimizing a distribution matching loss and an imitation loss on a composite policy, and uses such maps to transform expert's demonstration to what can be directly imitate in target domain.
However, this requires additional data of the corresponding demonstration for the same task in the two domains. 
Dripta et al. \cite{raychaudhuri2021cross} extends DAIL to a LfO setting, but did not get rid of the dependence on additional data.

By training two classifiers to capture the domain difference between the source domain and the target domain, Domain Adaptation with Rewards from Classifiers (DARC) \cite{eysenbach2020off} can train a near-optimal policies for the target domain on source domain.
We get inspiration from it, using the same method to capture domain differences, and introducing them into imitation learning to solve our problems.

\section{Background}

Our framework mainly considers two MDPs: $\mathcal{M}_{\text{source}}$ and $\mathcal{M}_{\text{target}}$. 
The former corresponds to the source domain which the demonstration is obtained, and the latter refers to the target domain where the reward function is desired.
A typical example of this setting is that the agent may be required to learn an action having been demonstrated in the real world (source domain), and the learning process is conducted in simulation (target domain). 
In the above scenario, the agent can interact in the target domain limitlessly, and the interaction in the source domain should be limited to a lower frequency.
We assume that two domains have the same state space $\mathcal{S}$, action space $\mathcal{A}$, and initial state distribution $p(s_0)$. 
The only difference between them is the transition dynamics: $p_{\text{source}}(s_{t+1} | s_t, a_t)$ and $p_{\text{target}}(s_{t+1} | s_t, a_t)$. 



\subsection{Adversarial inverse reinforcement learning}


This sub-section briefly describs our baseline, the AIRL \cite{fu2017learning} algorithm, which shares a similar skeleton with our method but differs in purpose and problem setting. 
AIRL is a model-free adversarial learning framework rooted from GAN, where the policy $\pi$ learns to imitate the behavior of the expert policy $\pi_E$ by minimizing the Jensen-Shannon divergence between the state-action distributions yielded by the two policies respectively. 
The learning process is achieved with the following objective:
\begin{scriptsize}
    \begin{equation}\label{form of GAN}
        \begin{split}
            \min _{\pi} \max_{{D \in(0,1)^{\mathcal{S} \times \mathcal{A}}}} 
            \{\mathbb{E}_{\pi_{E}}[\log D(s, a)]+
            \mathbb{E}_{\pi}[\log (1-D(s, a))]-\lambda H(\pi)\} \text{,}
        \end{split}
    \end{equation}
\end{scriptsize}
where $D$ is the discriminator that performs binary classification to distinguish if the sampled state-action pairs are generated by $\pi$. 
$H(\pi)$ is an entropy regularization term, and $\lambda$ is a hyper-parameter to scale it.  
AIRL concurrently unveils the reward function and improves the policy using the following discriminator:
\begin{small}
    \begin{equation}
        \label{airl_discriminator}
        D_{\psi}\left(s_t, a_t\right)=\frac{\exp \left[f_{\psi}\left(s_t, a_t\right)\right]}{\exp \left[f_{\psi}\left(s_t, a_t\right)\right]+\pi(a \mid s)} \text{.}
    \end{equation}
\end{small}
And the reward term $f_{\psi}(s_t, a_t)$ can break down into:
\begin{small}
    \begin{equation}
        \label{AIRL's reward}
        f_{\xi,\varphi}(s_t, a_t, s_{t+1})= g_\xi(s_t, a_t) + \gamma h_\varphi(s_{t+1}) - h_\varphi(s_t) \text{.}
    \end{equation}
\end{small}
As shown above, the reward function comprises a disentangled reward term $g_\xi(s_t, a_t)$ parameterized by $\xi$, and a shaping term $F = \gamma h\varphi(s_{t+1}) - h\varphi(s_t)$ parameterized by $\varphi$. 
Define $\psi = (\xi,\varphi)$. 
As described in AIRL, $g_\xi(s_t)$ can be parametrized as solely a function of the state, which we then use in the experiment for the reward function visualization.
The entire $D_{\psi}(s_t, a_t)$ is trained as a binary classifier to distinguish between expert demonstrations $\tau_E$ and policy generated trajectories $\tau$. 
The policy is trained to maximize the discriminative reward $\hat{r}(s_t, a_t) = \log(1-D(s_t, a_t)) - \log(D(s_t, a_t))$. 

\section{Off-dynamics Inverse Reinforcement Learning}





However, since AIRL requires the learning agent to interact with the same domain as the demonstration is in, it cannot be applied to our scenario. 
We expect to imitate real-world demonstrations by learning from interactions with the simulation environment, within the setting described above. 
In this paper, we aim to propose a solution to realize IRL from Hetero-Domain, the meaning and goal of which can be articulated as below:
\begin{Def}
    \textbf{IRL from Hetero-Domain} is the problem aiming to acquire a reward function in $\mathcal{M}_{target}$ by imitating demonstrations from $\mathcal{M}_{source}$. 
    The number of interactions in $\mathcal{M}_{source}$ should be limited, since it represents the real environment that has been demonstrated by experts.
    \label{prob_set}
\end{Def}
To solve this problem, we present our ODIRL method. 
First, we define a surrogate expert policy $\pi^{t}_{E}$ denoting expert in $\mathcal{M}_{\text{target}}$, which in fact does not exist. 
It represents the policies that should be adopted in the target domain to complete tasks consistent with the source domain experts’ demonstration.
The objective of the discriminator is also to minimize cross-entropy loss between the demonstrations from $\pi^{t}_{E}$ and the policy generated samples:
\begin{footnotesize}
    \begin{equation}\label{objective}
        \begin{split}
            \max_{\psi} 
            \{\mathbb{E}_{(s, a)\sim\pi^{t}_{E}}[\log D_{\psi}(s, a)]+ 
            \mathbb{E}_{(s, a)\sim\pi}[\log (1-D_{\psi}(s, a)]\} \text{.}
        \end{split}
    \end{equation}
\end{footnotesize}
This is a standard imitation learning optimization objective in the target domain. 
The form of $D_{\psi}(s, a)$ is same as Eq.\eqref{airl_discriminator}.
Combined with the policy optimization objective as the following reward $\hat{r}(s, a)=\log \left(1-D_{\psi}(s, a)\right)-\log \left(D_{\psi}(s, a)\right)$.
$\exp [f_{\psi}(s_t, a_t)]$ would capture actual distribution of $(s, a)\sim\pi^{t}_{E}$ when the discriminator is optimal, and $\pi$ would capture $\pi^{t}_{E}$ when maximize the $\hat{r}(s, a)$.
We define $r$ as the true reward functions of $\pi^{s}_{E}$ to be maximized in the source, and $f_{\psi}$ is learned to approximate the true reward functions of $\pi^{t}_{E}$  to be maximized in the target domains.
Then we have
\begin{small}
    \begin{equation}
        p_{\pi^{s}_{E}}(\tau) \propto p(s_{0}) \prod_{t=0}^{T} p_{\text{source}}(s_{t+1} \mid s_{t}, a_{t}) \exp(r(s_{t}, a_{t}))
    \end{equation}
\end{small}
and
\begin{small}
    \begin{equation}
        p_{\pi^{t}_{E}}(\tau) \propto p(s_{0}) \prod_{t=0}^{T} p_{\text{target}}(s_{t+1} \mid s_{t}, a_{t}) \exp(f_{\psi}(s_{t}, a_{t})) \text{.}
    \end{equation}
\end{small}
In order to estimate the first expectation in \eqref{objective}, we constrain $\psi$ by minimizing the reverse KL divergence between these two distributions:
\begin{footnotesize}
    \begin{equation}
        \begin{split}
            &D_{\mathrm{KL}}( p_{\pi^{s}_{E}}(\tau) \| p_{\pi^{t}_{E}}(\tau) ) \\
            = \quad&\mathbb{E}_{\tau\sim\pi^{s}_{E}}[ \sum_{t=0}^{T} r(s_{t}, a_{t})] - \\
            &\mathbb{E}_{\tau\sim\pi^{t}_{E}}[ \sum_{t=0}^{T} f_{\psi}(s_{t}, a_{t}) + \mathcal{DD}(s_t,a_t,s_{t+1}) ] \text{,}
        \end{split}
    \end{equation}
\end{footnotesize}
where the dynamics difference, $\mathcal{DD}(s_t,a_t,s_{t+1})$, is defined as:
\begin{footnotesize}
    \begin{equation}\label{dynamics difference}
        \begin{split}
            \mathcal{DD}(s_t,a_t,s_{t+1})\triangleq\log p_{\text {target}}(s_{t+1} \mid s_t,a_t)
            -\log p_{\text {source}}(s_{t+1} \mid s_t,a_t) \text{.}
        \end{split}
    \end{equation}
\end{footnotesize}
Then the \eqref{objective} could be modified to: 
\begin{footnotesize}
    \begin{equation}\label{ModifiedObjective}
        \begin{split}
            \max_{\psi} 
            \{ \mathbb{E}_{(s, a)\sim\pi^{s}_{E}}[\log D^{\prime}_{\psi}(s,a,s^{\prime})]+ 
             \mathbb{E}_{(s, a)\sim\pi}[\log (1-D_{\psi}(s, a)]\} \text{.}
        \end{split}
    \end{equation}
\end{footnotesize}
And the $D^{\prime}_{\psi}(s,a,s^{\prime})$ here is the discriminator which is modified by changing $r(s_{t}, a_{t})$ to $ f_{\psi}(s_{t}, a_{t}) + \mathcal{DD}(s_t,a_t,s_{t+1})$:
\begin{small}
    \begin{equation} \label{discriminator2}
        \begin{split}
            D^{\prime}_{\psi}\left(s_t, a_t, s_{t+1}\right)=\frac{\exp [f_{\psi}(\cdot)+\mathcal{DD}(s_t,a_t,s_{t+1})]}
            {\exp [f_{\psi}(\cdot)+\mathcal{DD} (s_t,a_t,s_{t+1})] + \pi(\cdot)} \text{,}
        \end{split}
    \end{equation}
\end{small}
when the data of demonstrations is fed in. 
For conciseness, $f_{\psi }(\cdot)$ and $\pi(\cdot)$ are used to denote $f_{\psi }\left(s_t, a_t \right)$ and $\pi\left(\mathbf{a}_{t} \mid \mathbf{s}_{t}\right)$, respectively. 

In the rest of this section, we will succeed the discussion of the calculation of $\mathcal{DD}(s_t,a_t,s_{t+1})$. 
Since its accurate value cannot be directly obtained, we approximate it with neural networks based on derivation which breaks down the value of $\mathcal{DD}(s_t,a_t,s_{t+1})$ into two binary classifiers.
Moreover, in the actual implementation, we multiply the estimated $\mathcal{DD}(s_t,a_t,s_{t+1})$ with a hyperparameter, $\alpha$, to scale its impact. 
Then, we carry out a brief explanation for our algorithm, including necessary elaborations on specific steps. 
\setlength{\textfloatsep}{5pt}
\begin{algorithm*}[t]
    \caption{ODIRL}
    \label{alg:example}
    \begin{algorithmic}[1]
        \STATE {\bfseries Input:} expert trajectories $\tau_{ex}$; expert replay buffer $\mathcal{B}_{ex}$; source MDP $\mathcal{M}_{source}$ and target MDP $\mathcal{M}_{target}$; ratio $r$ of experience from source vs. target; hyperparameter $\alpha$.
        \STATE {\bfseries Initialize:} replay buffers for source and target transitions, $\mathcal{B}_{source}$, $\mathcal{B}_{target}$; parameters $\theta = (\theta_{SAS}, \theta_{SA})$ for classifiers $q_{\theta_{SAS}}(\text{target} | s_t, a_t, s_{t+1})$ and $q_{\theta_{SA}}(\text{target} | s_t, a_t)$; policy $\pi$; discriminator $D_{\psi}\left(s_t, a_t, s_{t+1}\right)$
        \FOR {step $t$ in $\{1,\dots, N\}$}
            \STATE $\mathcal{B}_{target}\leftarrow\mathcal{B}_{target}\cup\mathrm{ROLLOUT}(\pi, \mathcal{M}_{target})$
                \IF{$t \mod r =0$}
                    \STATE $\mathcal{B}_{source}\leftarrow\mathcal{B}_{source}\cup\mathrm{ROLLOUT}(\pi, \mathcal{M}_{source})$
                \ENDIF
            \STATE Train $\theta \leftarrow \theta-\eta \nabla_{\theta} \ell(\theta)$ with $\mathcal{B}_{source}$ and $\mathcal{B}_{target}$
            \STATE Compute $\mathcal{DD}_{\theta}\left(s_{t}, a_{t}, s_{t+1}\right)$
            \STATE Optimize $\psi$ in Eq.\eqref{ModifiedObjective} via binary logistic regression to distinguish expert data $\mathcal{B}_{ex}$ from samples $\mathcal{B}_{target}$. 
            \STATE $\pi \leftarrow$ MAXENT $\operatorname{RL}\left(\pi, \mathcal{B}_{target}, f_{\psi}\left(s_t, a_t\right)\right)$
        \ENDFOR
    \end{algorithmic}
\end{algorithm*}

In practice, direct calculation of $\mathcal{DD}(s_t,a_t,s_{t+1})$ is technically unfeasible. 
We draw on \cite{eysenbach2020off}'s methodology which calculates the domain shift, since the domain shift quantification in \cite{eysenbach2020off} bears resemblance with $\mathcal{DD}(s_t,a_t,s_{t+1})$ in form. 

To be specific, $\mathcal{DD}(s_t,a_t,s_{t+1})$ can be further expanded as:
\begin{footnotesize}
    \begin{equation}\label{DD final form}
        \begin{split}
            \mathcal{DD}\left(s_{t}, a_{t}, s_{t+1}\right)
            &=\log p\left(\operatorname{target} \mid s_{t}, a_{t}, s_{t+1}\right)
            -\log p\left(\operatorname{target} \mid s_{t}, a_{t}\right)\\
            &-\log p\left(\text {source} \mid s_{t}, a_{t}, s_{t+1}\right)
            +\log p\left(\text {source} \mid s_{t}, a_{t}\right) \text{.}
        \end{split}
    \end{equation}
\end{footnotesize}
We learn two classifiers $q_{\theta_{SAS}} (\text{target} \mid s_t, a_t, s_{t+1})$ and $q_{\theta_{SA}} (\text{target} \mid s_t, a_t)$, parametrized by $\theta_{SAS}$ and $\theta_{SA}$ respectively.
And Eq.\eqref{DD final form} can be parameterized into a linear combination of the difference in logits from such classifiers.
The classifiers can be learned through the training process, and conducted by minimizing the standard cross-entropy losses:
\begin{small}
    \begin{equation*}
        \begin{split}
            \ell_{\mathrm{SAS}}\left(\theta_{\mathrm{SAS}}\right) \triangleq
            &-\mathbb{E}_{\mathcal{M}_{\mathrm{target}}}\left[\log q_{\theta_{\mathrm{SAS}}}\left(\text{target}\mid s_{t}, a_{t}, s_{t+1}\right)\right]\\
            &-\mathbb{E}_{\mathcal{M}_{\mathrm{source}}}\left[\log q_{\theta_{\mathrm{SAS}}}\left(\text{source}\mid s_{t}, a_{t}, s_{t+1}\right)\right]\\
            \ell_{\mathrm{SA}}\left(\theta_{\mathrm{SA}}\right) \triangleq
            &-\mathbb{E}_{\mathcal{M}_{\mathrm{target}}}\left[\log q_{\theta_{\mathrm{SA}}}\left(\text{target}\mid s_{t}, a_{t}\right)\right]\\
            &-\mathbb{E}_{\mathcal{M}_{\mathrm{source}}}\left[\log q_{\theta_{\mathrm{SA}}}\left(\text{source}\mid s_{t}, a_{t}\right)\right] \text{.}
        \end{split}
    \end{equation*}
\end{small}

To simplify notation, we can define $\theta=(\theta_{SAS}; \theta_{SA})$ and $\ell(\theta)=\ell _{SAS}(\theta _{SAS}) +\ell_{SA}(\theta _{SA})$. 
The training process of the two classifiers in Eq.\eqref{DD final form} will yield the value for the parameterized $\mathcal{DD}(s_t,a_t,s_{t+1})$, denoted as $\mathcal{DD}_\theta(s_t,a_t,s_{t+1})$.

\begin{figure}[t]
    \centering
    \includegraphics[width=0.26\textwidth]{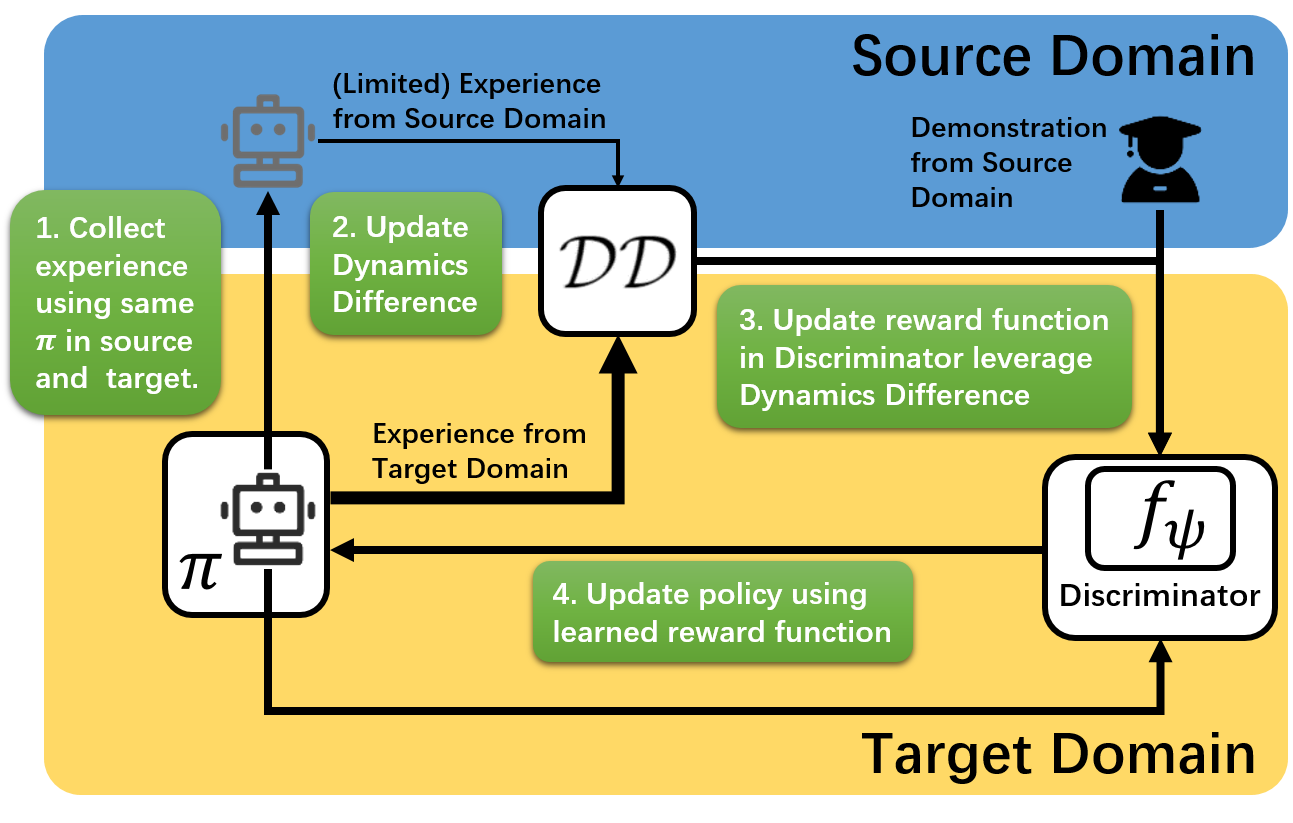}
    \vspace{-5pt}
    \caption{Block diagram of ODIRL (Alg.\ref{alg:example})}
    \label{diag2}
    \vspace{-10pt}
\end{figure}

Our entire algorithm for ODIRL is presented in Alg.\ref{alg:example} and illustrated in Fig.\ref{diag2}. 
In each iteration, we collect transitions from the source and (less frequently) target domain, and store the transitions in separate replay buffers. 
We then sample from both buffers to update the classifiers composing $\mathcal{DD}_{\theta}(s_t,a_t,s_{t+1})$. 
The classifiers are then leveraged to calculate the modified reward function in the source domain.
Then we train the discriminator, which reveals learned reward and its generator which represents optimal policy. 
With the assistance of $\mathcal{DD}_{\theta}(s_t,a_t,s_{t+1})$, when Alg.\ref{alg:example} is trained to convergence, $f_{\psi}\left(s_t, a_t\right)$ is the learned reward function in the target domain which is related to the current state and action. 
And $\pi$ minimizing the generator's loss reveals the optimal policy in the target domain. 
DRL can then be performed by utilizing the reward function $f_{\psi}\left(s_t, a_t\right)$ and generate a policy in the target domain. 
In real-life application scenarios, the policy will be transferred to the real world, which is beyond discussion of this paper.

\section{Experiments} \label{experiments}

\begin{figure}[ht]
    \centering
    \subfigure[]{
        \includegraphics[width=0.07\textwidth]{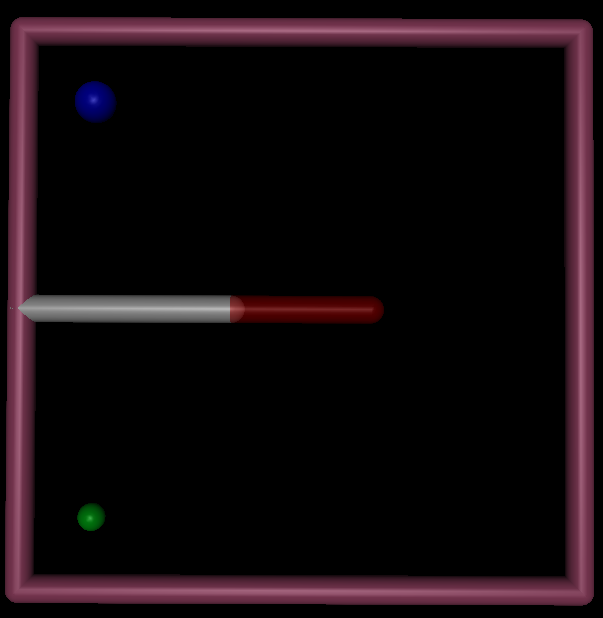}
        \label{env1}
        }
    \subfigure[]{
        \includegraphics[width=0.07\textwidth]{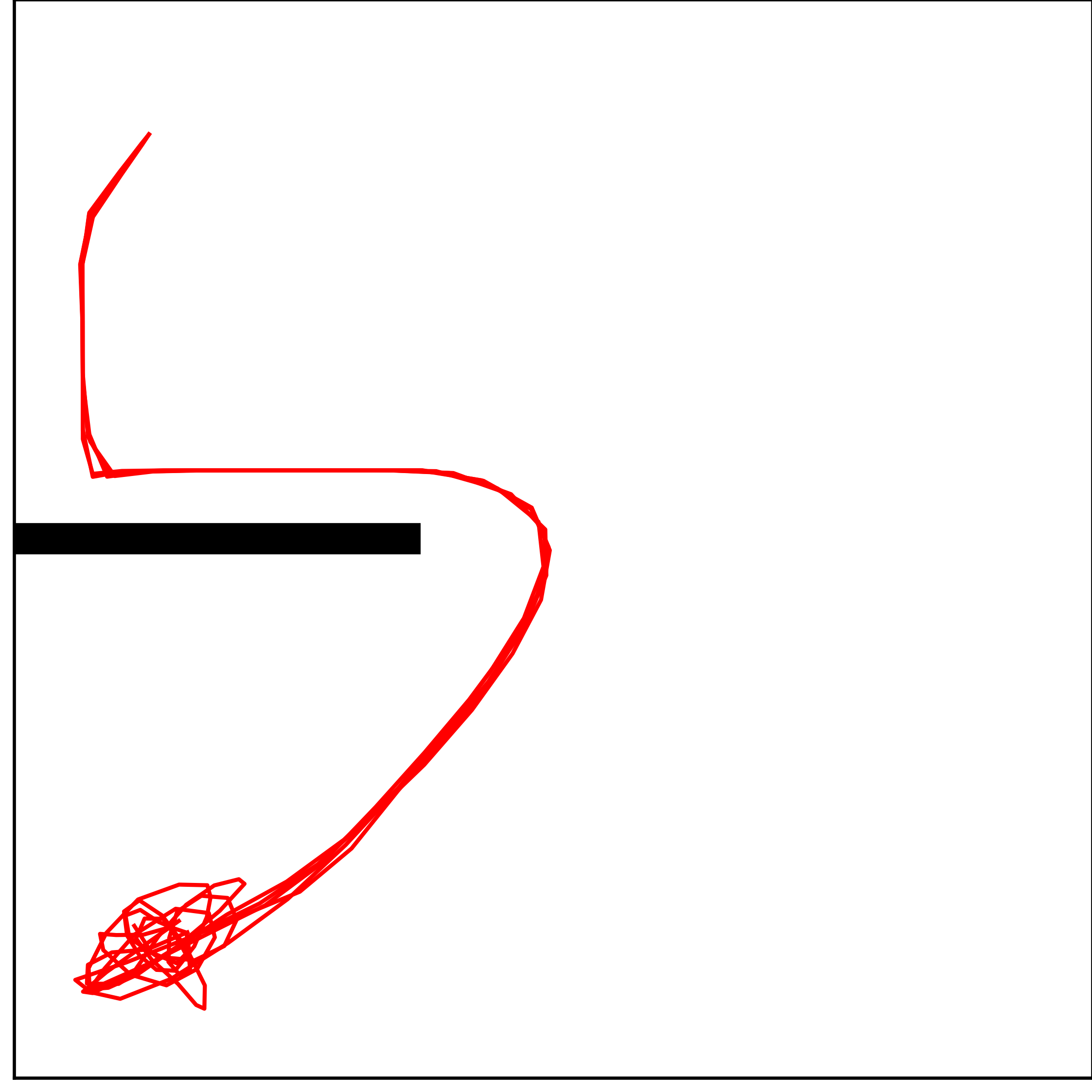}
        \label{exp1:demo}
    }
    \subfigure[]{
        \includegraphics[width=0.07\textwidth]{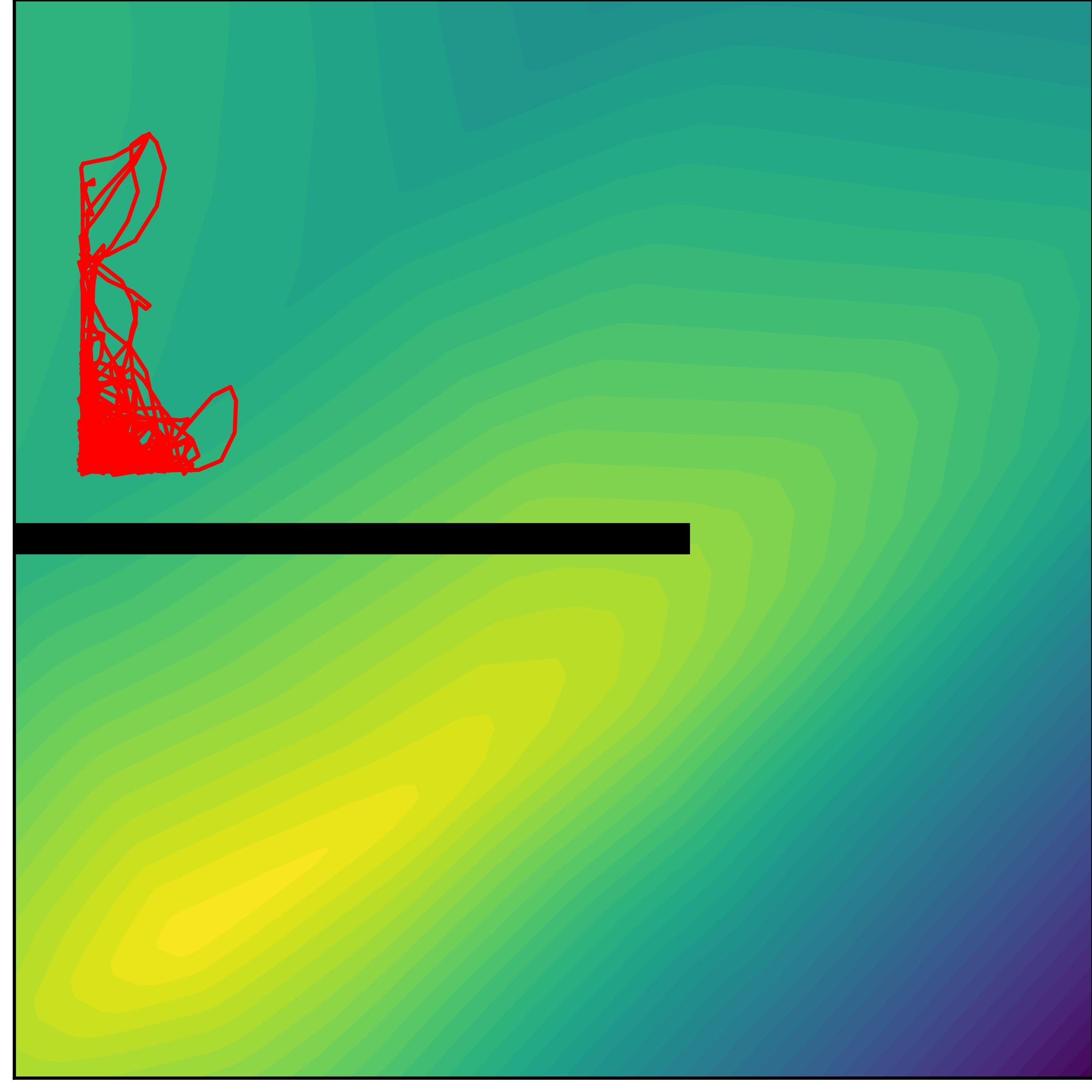}
        \label{exp1:airl}
    }
    \subfigure[]{
        \includegraphics[width=0.07\textwidth]{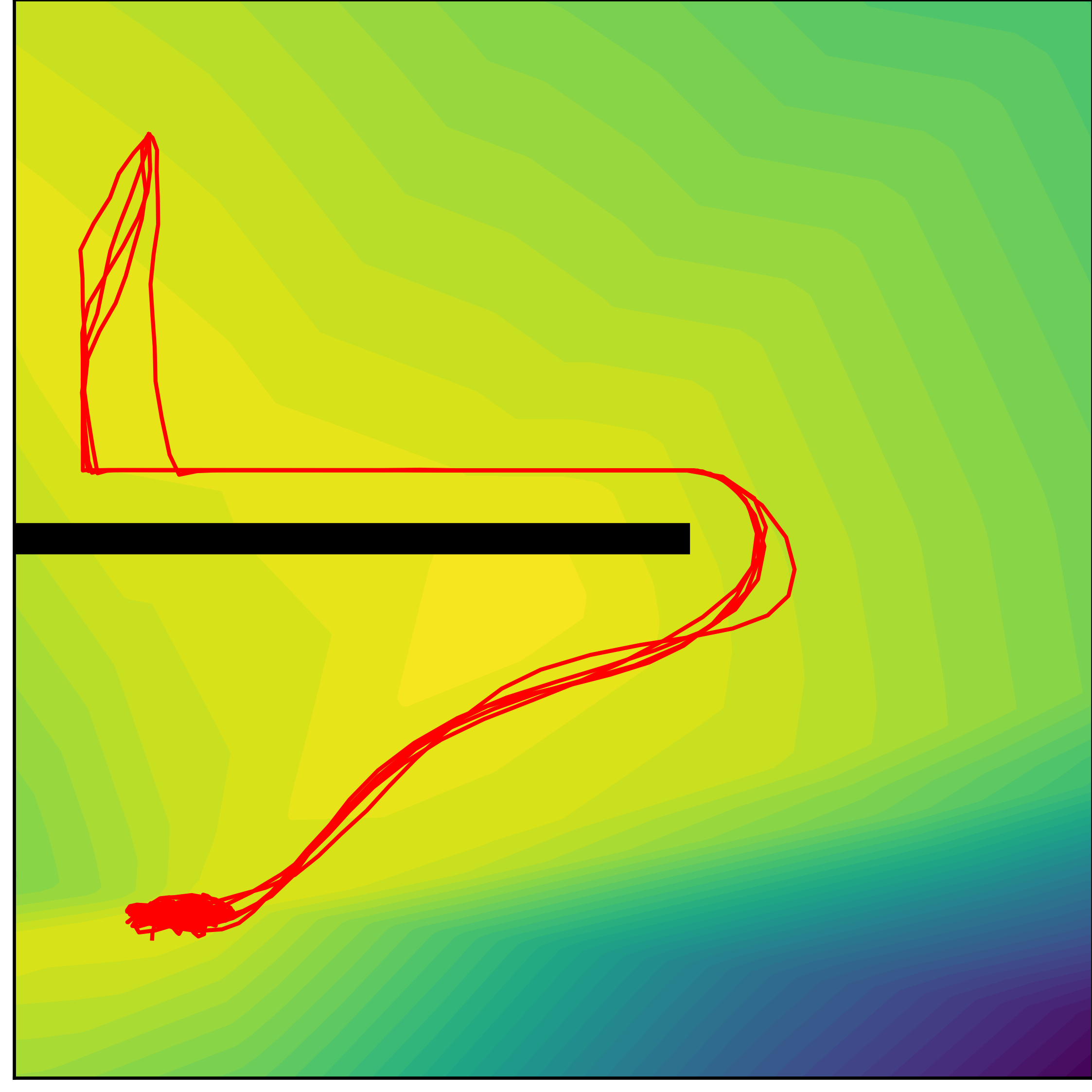}
        \label{exp1:odirl}
    }    
    \label{exp1}
    \caption{ \textbf{(a)} The Point-Maze environment \textbf{(b)} Expert's trajectories given in the source domain. \textbf{(c)\&(d)} Outcome of AIRL and ODIRL.
    }
    \vspace{-10pt}
\end{figure}

\textbf{Toy example.}
We start with a simple MuJoCo PointMaze example shown in Fig.\ref{env1}. 
In the source domain, the agent demonstrates a task to navigate itself from the blue point at the top left corner to the green point at the bottom left corner. 
To achieve this, it has to get around an obstacle. We train the expert using the Soft Actor-Critic (SAC) algorithm \cite{haarnoja2018soft}.
In the target domain, the obstacle is wider, and the odds in obstacle width is marked with the translucent red bar in Fig.\ref{env1}.
Such setting results in the difference of dynamics in the two domains.
In this scenario, given the obstacle-circumventing demonstration in the source domain shown in Fig.\ref{exp1:demo}, we aim to acquire a reward function to guide accomplishment of a similar task in the target domain. 
If we simply apply the AIRL on the target domain, the policy generated from the obtained reward function fails the task as predicted since the reward function does not capture the domain discrepancies. 
Fig.\ref{exp1:airl} shows that the agent trained with reward functions (shown in heatmap) yields by AIRL lingers around at the corner which it considers as the most rewarding position since it is closest to the destination. 
On the contrary, our method successfully enables the agent to achieve the designated task, which is shown in Fig.\ref{exp1:odirl}. 

\begin{figure*}[t] 
    \centering
    \includegraphics[width=0.98\textwidth]{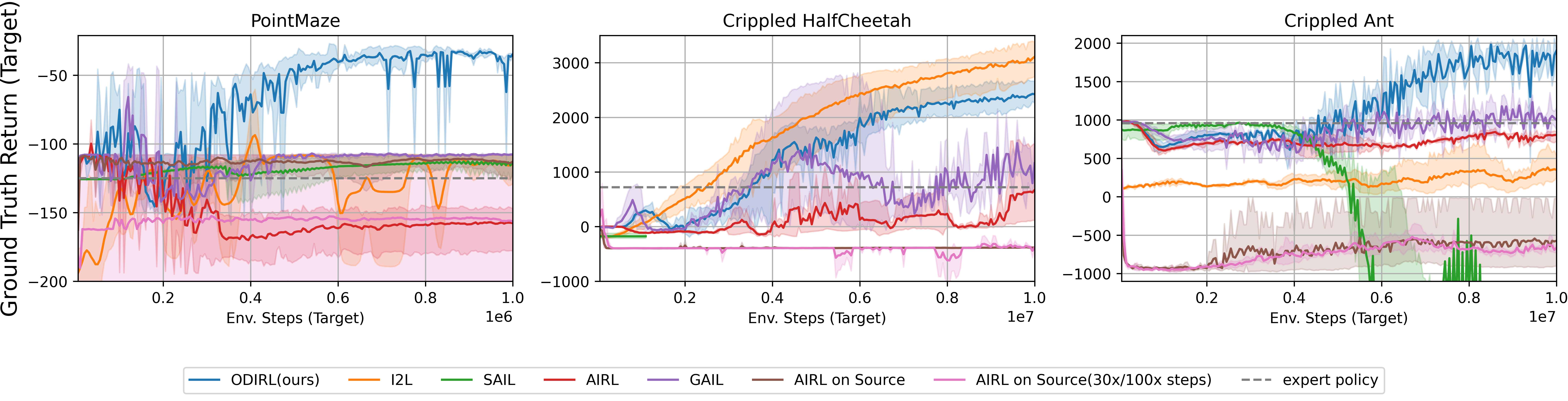}
    \caption{\textbf{Training Process of ODIRL and Baselines on Three Tasks}. The dynamics differences are created by widening the obstacle in the PointMaze task and disabling joints in the Half Cheetah task and the Ant task.
    The gray dashed curves present the returns yielded by directly executing the expert policy in the target domain.}
    \label{acc}
    \vspace{-20pt}
\end{figure*}


\textbf{Scaling to more complex tasks.}
We now demonstrate the effectiveness of ODIRL on more difficult tasks. 
The algorithm is tested with two types of simulation robots in the OpenAI Gym environment \cite{brockman2016openai}, the Half Cheetah and the ant.  
To create dynamic differences in the two domains, we use fully functional robots in the source domain for demonstration, whereas the robots used in the target domain have several joints disabled. 
The latter is to imitate the demonstrations given by the former, which, in both environments, are the motion of running forward. 
Moreover, the crippled joints in the source domain are circled in red in both figures. 
To be specific, on the Half Cheetah robot, the shoulder joint and the hip joint are disabled; on the ant robot, the four joints on the hind legs are disabled. 


The performance of our method is compared with that of other six baselines. 
The \textbf{expert policy} is trained with SAC performed on the source domain, and is executed in the same domain to generate demonstration.
The \textbf{expert policy} is also directly utilized in the target domain for comparison, the resulting returns of which are shown in the gray dashed curves in Fig. \ref{acc}. 
\textbf{I2L} \cite{Gangwani2020State-only} and \textbf{SAIL} \cite{liu2019state} also train in the target domain, by imitating the state trajectories in expert's demonstration.
\textbf{AIRL} and \textbf{GAIL}, are trained in the target domain with demonstrations given in the source domain, which is the same as the setting of our own method. 
Note that these two baseline methods both originally require the learning agent to interact with the domain where demonstrations are provided. 
Here we simply use the SAC-generated demonstration from source. 
This is feasible technically because the demonstration is, after all, an sequential array of state-action pairs.
However, such maneuver makes no sense as the demonstration may include impractical or unsuitable behavior in the target domain. 
We operate like this just to create a meaningful baseline for discussion.
We also directly apply AIRL on the source domain and transfer the disentangled reward $g_\xi(s_t)$ (shown in Eq.\eqref{AIRL's reward}) to the target domain, which conforms to the standard setting of AIRL. 
Recall that ODIRL collects much more transition records in the target domain than in the source domain, which is caused by the hyperparameter $r$ (see Alg.\ref{alg:example} Line 5). 
$r$ denotes the ratio of target/source domain interaction recordings.
For fairness, we also compare our method with AIRL having gone through the same number of steps of gradient updates for each single collected transition in the source domain. 
In practice, we increase the gradient update times to 30/100 times. 
In more detail, the number of 30 and 100 corresponds to the $r$ value which we adopt for different tasks. 
For the PointMaze task, $r$ is 30. For the other two tasks, $r$ is 100.


\begin{figure}[ht] 
    \centering
    \includegraphics[width=0.35\textwidth]{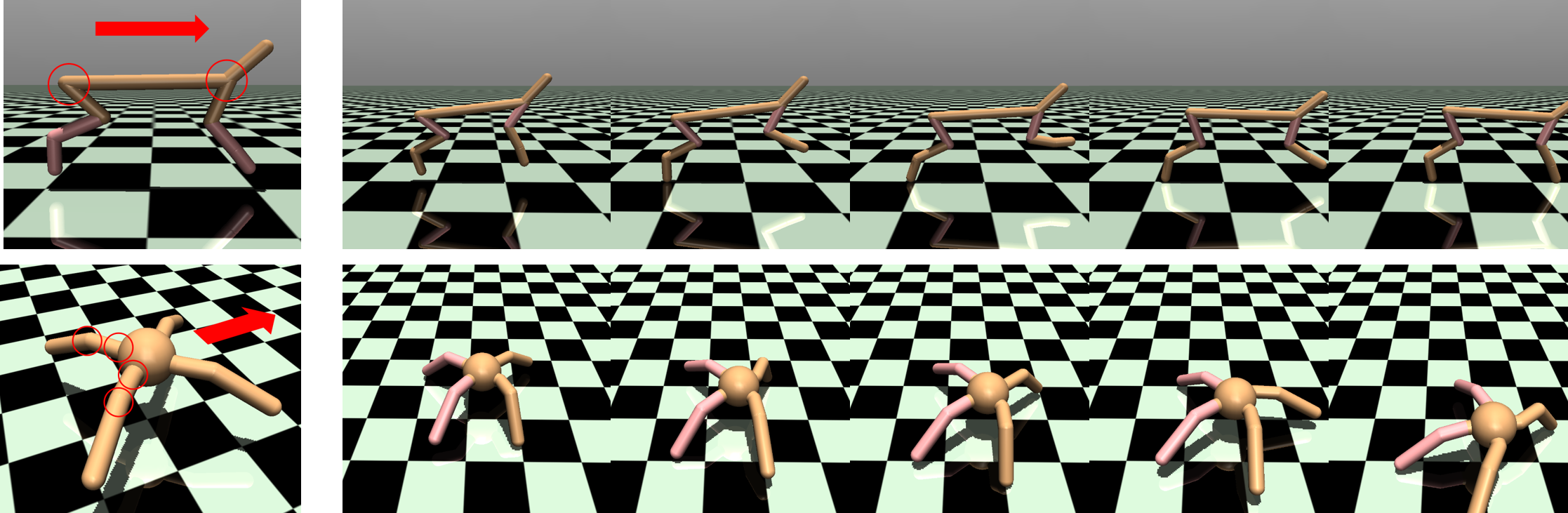}
    \caption{ Demonstration of the motion produced by the policy learned with the reward function acquired with ODIRL in Half Cheetah and Ant tasks.}
    \label{ant}
    \vspace{-10pt}
\end{figure}


The results of the experiments are shown in Fig.\ref{acc}. 
The main criterion we refer to for comparison is based on the ground truth reward function, and its cumulative sum over episodes, named as the ground truth return. 
The ground truth reward function is simply the negative value of the distance between the agent and the goal.
We evaluate each method by analyzing the ranges of the ground truth returns yielded by three random seeds with which the methods and the environments are initialized. 
The ranges are represented by the shaded area in Fig.\ref{acc}, and the mean values of them for each method are marked with thick curves. 

For all tasks, the expert policies trained in the source domain but directly used in the target domain yield poor performance, as indicated by the gray dashed straight lines in Fig.\ref{acc}. 
Such result coincides with expectation since dynamics discrepancies between the two domains entails distinct action strategies. 
Rotely copying of the policy is likely to be in vain.
Note that the dashed lines do not reflect the return changes in the training process because the policies are not trained in the target domain.

As for the \textbf{AIRL on source} method and the \textbf{AIRL on source (30x/100x steps)} method, in principal, they are targeted at the acquisition of a transferable reward function, which seems like a closer solution for the domain adaptation problem.
We align the setting of their training with that of our method with a purpose to equalize the amount of information which these methods can obtain from the two domains.
To be specific, we run same steps in the source-domain for the two AIRL-based baselines and our method, and the training steps of the all three methods in the target domain are the same too.
It is worth mentioning that all the policies are trained with the Proximal Policy Optimization (PPO) algorithm \cite{schulman2017proximal} in the target domain, including our method and the compared baselines.
As shown in Fig.\ref{acc}, \textbf{ODIRL} beats all baselines in most instances. 
In the environment of PointMaze and Ant, may due to the large dynamic difference between the source domain and the target domain, the performance of our method far exceeds the baseline. 
In the Half Cheetah environment, our method also has comparable performance. 
With only a limited quantity of interaction with the source domain, the two AIRL-based baselines cannot produce informative and accurate reward functions in the first place, not to mention to generate a well-functioning policy with those reward functions, even these two methods can afterwards endow the learning agent with a comparable amount of information in the target domain to that of our method.
On the other hand, \textbf{ODIRL}, organically integrating the information acquisition process in the source domain and in the target domain, can achieve a significantly better performance in all the three tasks.

\textbf{AIRL} and \textbf{GAIL} are relatively similar methods, except that the former explicitly reveals the reward function and trains the generator with it. 
Surprisingly, although \textbf{GAIL} does not consider transferability in the process of imitation, it outperforms \textbf{AIRL}. This may indicate that the dynamics difference between the two domains may turn the intention of \textbf{AIRL} into a backfire. 
After all, the disentangled reward contains only state-only information, which may be distorted in the presence of dynamics difference.     
Yet, our method generally defeats these two method with better outcomes and smaller variances.

\begin{figure}[t]
    \centering
    \includegraphics[width=0.4\textwidth]{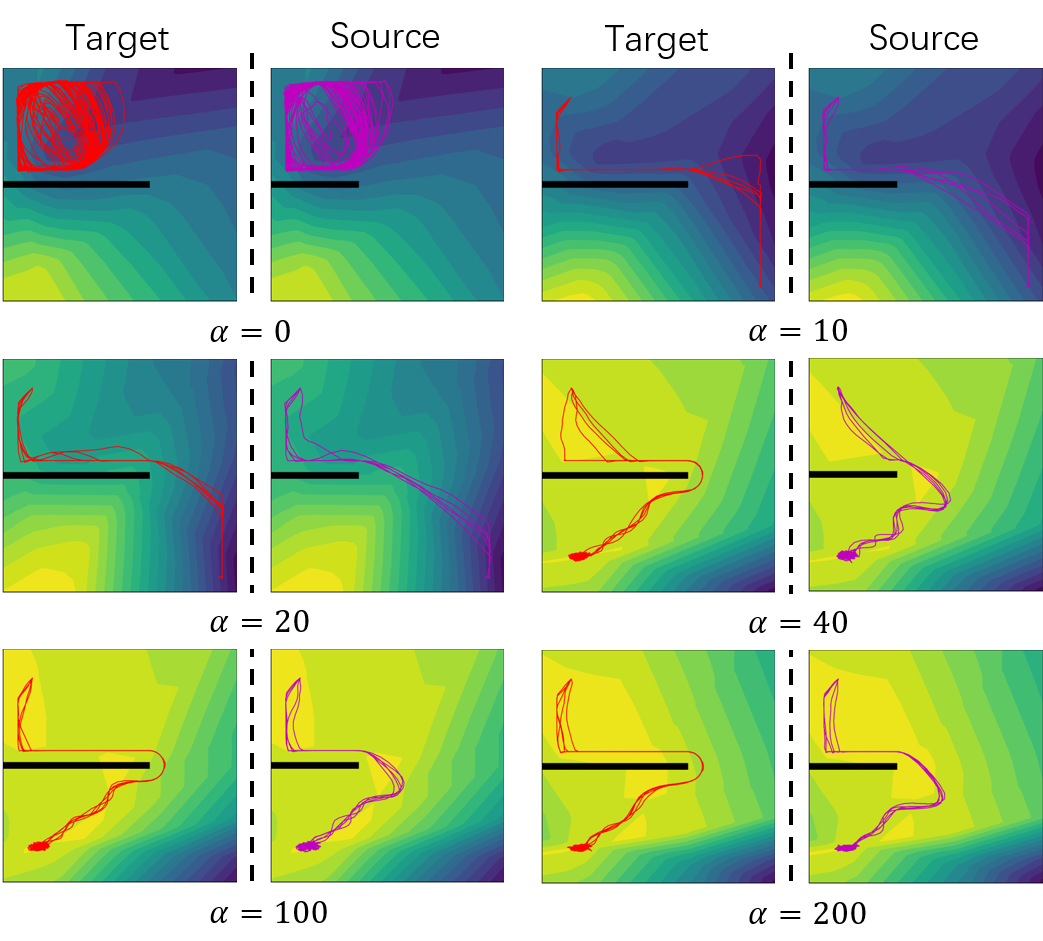}
    \caption{Results of ablation experiments}
    \label{ablation}
    \vspace{-10pt}
\end{figure}


\textbf{Ablation analysis.} 
To gain more intuition for our method, we alter the hyperparameter, $\alpha$, in the PointMaze experiment to demonstrate how $\mathcal{DD}(s_t,a_t,s_{t+1})$ functions in the algorithm. 
$\alpha$ scales the impact of $\mathcal{DD}(s_t,a_t,s_{t+1})$ in Eq.\ref{discriminator2}.
In effect, this hyperparameter decides how much the discriminator refrains from rewarding demonstrations that exploit the discrepancies between the two domains. 
In other words, if $\alpha$ equals to $0$, then ODIRL falls back to AIRL, which merely imitates the trajectories of the demonstration.
However, the transition dynamics disparities between the two domains may cause the imitation to fail, as will be shown in the following example. 
As $\alpha$ takes higher value in our framework, the dynamics difference is increasingly involved in the calculation of the discriminator. 
The discriminator will then make more effort to maximize the similarity between the trajectories resulted by the learning agent's interaction with the two domains. 
The effects of different $\alpha$'s are shown in Fig.\ref{ablation}, which shows the state-only reward functions (shown in heatmap) and the trajectories yielded by the generator of ODIRL training under different values of $\alpha$. 
Increasing $\alpha$ helps learned reward function take dynamics different into consideration, and makes imitation learning easier. 
The trajectories in red are the ones executed with the policies trained with the acquired reward functions in the target domain, while the trajectories in purple are yielded by the same policies in the source domain. 
When the $\alpha$ is small, the generated reward functions fail to guide the learning agent to perform the expected motion, which proves that stiff imitation of demonstration without considering dynamics difference will fail the task.
When $\alpha$ is assigned a larger value, the reward function will be induced to a landscape which can generally achieve the imitation task while making necessary adjustments to overcome the dynamics difference, thus the domain adaptation is realized.


\section{Discussion}
In this paper, we propose a simple, practical, and intuitive approach for IRL from Hetero-Domain. 
We draw motivation from a novel variational perspective on IRL from Hetero-Domain, which suggests that we can add constraints so that the reward function can be obtained by imitating the demonstration in different dynamic domains.
Experiments on control tasks show that our method can leverage the demonstration from source domain to learn reward functions which perform well in the target domain, with observing only a small amount of transitions from the source domain.









\bibliography{ref}  
\bibliographystyle{IEEEtran}

\end{document}